\documentclass[11pt]{article}
\pdfoutput=1

\usepackage[final]{slai}

\usepackage[utf8]{inputenc}
\usepackage[T1]{fontenc}
\usepackage{mathpazo}
\usepackage[scaled=1.1]{zlmtt}
\usepackage{amsmath,amssymb,bm}
\usepackage{dsfont}
\usepackage{booktabs}
\usepackage{graphicx}
\usepackage{float}
\usepackage{listings}
\usepackage{microtype}
\usepackage{xcolor}
\usepackage{hyperref}
\usepackage[nameinlink,capitalise]{cleveref}

\graphicspath{{figures/}}
\hypersetup{
  colorlinks=true,
  linkcolor=blue!60!black,
  citecolor=blue!60!black,
  urlcolor=blue!60!black,
  pdftitle={AscendGYM: A Cost-Aware Kernel Optimization Benchmark for Real-World Workloads on Ascend NPUs},
  pdfauthor={AI Training Platform Team, Shenzhen Loop Area Institute}
}

\newcommand{\reporttitle}{AscendGYM: A Cost-Aware Kernel Optimization Benchmark for Real-World Workloads on Ascend NPUs}
\providecommand{\reportsubtitle}{}
\providecommand{\reportauthor}{}

\newcommand{\reportaffiliation}{AI Training Platform Team, Shenzhen Loop Area Institute\\}
\newcommand{\reportdate}{\today}

\title{\reporttitle\\[0.4em]\large\reportsubtitle}
\author{\reportauthor\\\reportaffiliation}
\date{\reportdate}

\begin{document}

\maketitle

\begingroup
\NoHyper
\renewcommand{\thefootnote}{}
\footnotetext{%
  \raggedright
  For any questions or feedback regarding this technical report, please contact Dongfang Li (\texttt{lidongfang@hit.edu.cn}) and Baotian Hu (\texttt{baotianhu@slai.edu.cn}).
}
\endNoHyper
\endgroup

\begin{abstract}
Kernel optimization is essential for efficient large language model (LLM) training and inference on Ascend NPUs.
Existing Ascend kernel benchmarks primarily focus on kernel generation, with less emphasis on the ability of agents to optimize existing implementations under input configurations drawn from real workloads.
To address this requirement, we introduce AscendGYM, a benchmark whose initial suite comprises 48 kernels across three difficulty levels.
Each kernel is paired with test cases whose data types and tensor shapes are collected from real-world training and inference tasks.
Consequently, performance improvements on AscendGYM map directly to the corresponding workloads. In terms of usability, two evaluation modes are supported in AscendGYM: model evaluation and model--harness evaluation. The kernel optimization capabilities of both trained models and custom agents are allowed to be measured. It provides a complete execution environment that closely reproduces how Ascend kernels are invoked in practical pipelines. For model evaluation, it additionally includes a built-in harness equipped with search orchestration and experience accumulation. Finally, we propose a new metric, the Speedup--Token Score (STS), which quantifies performance gains relative to token cost. By jointly considering speedup and cost, STS enables the assessment of practical kernel optimization capability of a model or harness under an acceptable budget. AscendGYM is available at \href{https://github.com/SLAI-AITP/AscendGYM}{https://github.com/SLAI-AITP/AscendGYM}.
\end{abstract}

% \tableofcontents
% \newpage

\section{Introduction}
\label{sec:introduction}

Efficient large language model (LLM) training and inference depend on the performance of the underlying kernels. On Ascend NPUs, improving kernel implementations for real workloads is an important engineering step toward greater system efficiency~\citep{li2026trex}. Such optimization typically starts from existing code and requires developers to analyze performance bottlenecks in the context of target input configurations and hardware characteristics, then repeatedly modify and execute implementations to validate improvements~\citep{andrews2025gpu}. The process therefore relies on specialized expertise and sustained engineering effort~\citep{tschand2025swizzleperf}.

Coding models and agents offer an opportunity to automate this process~\citep{baronio2026kevin}. Numerous benchmarks have been proposed to evaluate their capabilities. KernelBench evaluates the ability of models to generate efficient kernels for given PyTorch workloads~\citep{ouyang2025kernelbench}, while NPUKernelBench and CANN Bench extend kernel-generation evaluation to Ascend, assessing compilation, correctness, and performance~\citep{cao2026ascendkernelgen,gao2026cannbench}.
We focus on a concrete optimization evaluation task: given a correct kernel implementation, can the specific agent further improve its performance for a target workload within an acceptable budget while preserving correctness?
To study this task, we introduce \textbf{AscendGYM}, a kernel optimization benchmark for Ascend NPUs, whose design centers on the following questions:

\paragraph{How should evaluation tasks be designed?}
Optimization priorities for the same kernel can vary with input configurations~\citep{ye2025flashinfer}. For example, attention computations process different data sizes during prefill and decoding, which can favor different ways of organizing computation~\citep{kamath2025pod}. FlashInfer-Bench draws workloads from real inference traces, illustrating the connection between input configurations and application scenarios~\citep{xing2026flashinferbench}.
AscendGYM selects kernels and their input shapes, data types, and computation parameters from real training and inference workloads, and provides existing implementations together with corresponding test cases.
Each task thus has a concrete optimization target and defined validation conditions, with performance gains measured relative to the initial implementation's latency under the same conditions.

\paragraph{How can evaluation be made reliable?}
Whether modified code compiles, produces correct results, and runs faster must be checked separately through actual execution~\citep{li2025autotriton}.
Prior work shows that compiler and profiling feedback can help models correct errors and improve implementations~\citep{ouyang2025kernelbench}.
On Ascend, this process also involves coordinated builds of host and device code, followed by execution and measurement on NPUs~\citep{cao2026ascendkernelgen}.
We design \textbf{AscendForge}, AscendGYM's built-in execution sandbox, to integrate candidate compilation, correctness checking, and performance measurement into a programmatically accessible evaluation service.
It returns execution feedback under predefined validation conditions and measurement protocols, and records optimization trajectories to support iterative agent optimization and result verification.

\paragraph{How should optimization performance be assessed?}
Repeated sampling and iterative refinement can improve results for some models, while requiring additional inference and execution resources~\citep{ouyang2025kernelbench}.
Two systems may achieve the same final speedup yet differ substantially in the search effort required: one may identify improvements within a few attempts, whereas the other may require an extended search~\citep{cha2025multi}. Final speedup alone does not capture this distinction.
\textbf{Speedup--Token Score (STS)}, which combines performance gains within a given budget with cumulative token consumption at each improvement, is designed to assess optimization under resource constraints.
The metrics for performance, resource consumption, and correctness are also reported separately to distinguish the gains achieved, the resources invested, and whether candidate implementations pass validation.

AscendGYM currently comprises \textbf{48 kernel optimization tasks} across \textbf{three difficulty levels} and supports \textbf{two evaluation modes}: comparing models using a default harness, and evaluating model--harness combinations as complete systems. Multiple leading models paired with the commonly used harnesses are systematically evaluated on the \textbf{17 tasks in Batch~1}. Approximately $70\%$ of the evaluated combinations achieve geometric mean speedups below $1.2\times$, indicating that substantial improvements across a broad range of existing implementations remain challenging. Gain trajectories under different budgets, together with differences in performance, cost, and correctness, also show that final speedup alone does not fully characterize optimization behavior. These findings provide empirical support for the benchmark's task and metric design.

The main contributions of AscendGYM are:
\begin{enumerate}
    \item \textbf{AscendGYM} is a benchmark for Ascend NPUs that uses kernels and input configurations from real workloads to evaluate the ability of agents to optimize existing implementations.
    \item \textbf{STS} is the key metric of AscendGYM, combining performance gains with the token consumption required to obtain them to assess optimization under a given budget.
    \item \textbf{AscendForge} built in AscendGYM supports iterative agent optimization through unified interfaces for compilation, correctness checking, performance measurement, and feedback.
\end{enumerate}

\section{Background and Related Work}
\label{sec:background}

Existing kernel generation benchmarks primarily evaluate whether models can construct correct and efficient implementations from computational requirements.
KernelBench~\citep{ouyang2025kernelbench} specifies tasks using PyTorch programs and asks models to generate semantically equivalent GPU kernels.
TritonBench~\citep{li2025tritonbench} focuses on Triton kernel generation, while MultiKernelBench~\citep{wen2025multikernelbench} extends evaluation to multiple hardware platforms.
For Ascend, NPUKernelBench~\citep{cao2026ascendkernelgen} provides functional descriptions, input and output requirements, and code templates to be completed, and evaluates the correctness and performance of the resulting implementations.
CANN Bench~\citep{gao2026cannbench} provides operator specifications, test cases, reference implementations for correctness checking, and performance baselines, and assesses candidate kernels against hardware performance limits.
These benchmarks establish the computational requirements, correctness criteria, and performance targets that generated implementations need to satisfy.

Some evaluations start directly from existing kernel implementations and assess agents' ability to improve their performance.
For example, AMD evaluates optimization of existing Triton kernels on TritonBench-modified and ROCm-bench, requiring agents to improve execution performance on AMD GPUs while preserving correctness~\citep{wang2025geaktritonv2}.
These settings use existing kernel source code as the optimization target and measure performance improvements through speedup relative to the initial implementation.

PRAGMA~\citep{lei2025pragma} incorporates execution feedback and hardware profiling into an iterative reasoning loop.
AutoKernel~\citep{jaber2026autokernel} profiles models to prioritize bottleneck kernels, then validates and benchmarks successive Triton or CUDA C++ implementations.

Beyond the starting implementation and target platform, the relationship between kernels, input configurations, and actual applications also affects the practical relevance of performance evaluation.
SOL-ExecBench~\citep{lin2026solexecbench} extracts kernel tasks from real models and uses hardware performance limits to assess the remaining optimization headroom.
Atrex-Bench~\citep{yang2026atrexbench} and FlashInfer-Bench~\citep{xing2026flashinferbench} construct evaluation tasks from execution traces of real inference workloads, connecting kernels and their input configurations to specific deployment scenarios.
These efforts illustrate how workload-driven task design gives kernel performance evaluation a concrete application context.

AscendGYM focuses on evaluating the optimization of existing kernel implementations on Ascend NPUs.
Its tasks comprise kernels used in real large-model workloads, with input shapes extracted from the corresponding workloads to reflect concrete execution requirements.
Each task provides complete, runnable initial source code and requires agents to improve performance while preserving correctness, using the measured latency of the initial implementation under the same input configuration as the reference.
AscendGYM includes the AscendForge execution environment, which provides shared support for candidate compilation, correctness checking, and performance measurement.
Under explicit evaluation protocols and budgets, we compare the optimization performance of different models and harnesses, considering performance gains, correctness, and resource consumption.

\section{AscendGYM}
\label{sec:methodology}

AscendGYM is a kernel optimization benchmark for Ascend NPUs. It evaluates the ability of model--harness combinations to improve existing kernel implementations. Its four components are the task set and test cases, the AscendForge evaluation environment and tools, evaluation metrics, and a shared evaluation workflow. Together, they provide tasks and execution feedback and assess optimization outcomes under consistent rules.

\subsection{Task Selection and Test Case Design}
\label{sec:task-selection}
\label{sec:philosophy}

AscendGYM organizes kernel optimization tasks into three batches of increasing difficulty, progressing from largely independent kernels to multistage fusion and more complex execution coordination. Batch~1 focuses on memory access optimization for basic kernels. Batch~2 covers multistep computations, such as normalization and quantization. Batch~3 targets
complex attention mechanisms and fused computations, including sparse attention. This report evaluates the 17 kernel optimization tasks in Batch~1; results for Batches~2 (17 kernels) and~3 (14 kernels) will be reported in future updates.

Task selection prioritizes bottleneck kernels with substantial performance impact in real workloads. Each test case preserves the connection between kernel inputs and their deployment context, including the target model, execution phase, hardware and parallelism configurations, and the resulting tensor shapes, data types, and computational parameters. This context links
measured kernel speedups to the corresponding operator invocations in real workloads, aligning evaluation with deployment requirements.

For example, the \texttt{AddRMSNorm} task captures residual addition followed by RMS normalization during a forward pass of Qwen3-0.6B~\citep{yang2025qwen3} inference. The test case uses the prefill phase with no quantization, eager execution, a single NPU, TP~1, BF16, a batch size of~1, and an input sequence length of 32~tokens. The branch output and residual tensors both have shape $[32,1024]$, and the normalization weight has shape $[1024]$. Profiling of its workload shows 56 invocations per forward pass, accounting for 8.77\% of total computation time and ranking among the top-5 kernels by cumulative execution time. Preserving this execution context supports assessment of contribution to workload-level performance.

AscendGYM evaluates performance across multiple deployment cases, the best achievable performance for specific shapes and scenarios, and the optimization cost required to obtain these gains. Together, these dimensions characterize cross-case applicability, shape specialization, and optimization efficiency.

\subsection{Evaluation Environment and Tools}
\label{sec:ascendforge}

\begin{figure}[t]
  \centering
  \makebox[\textwidth][c]{\includegraphics[width=1\textwidth]{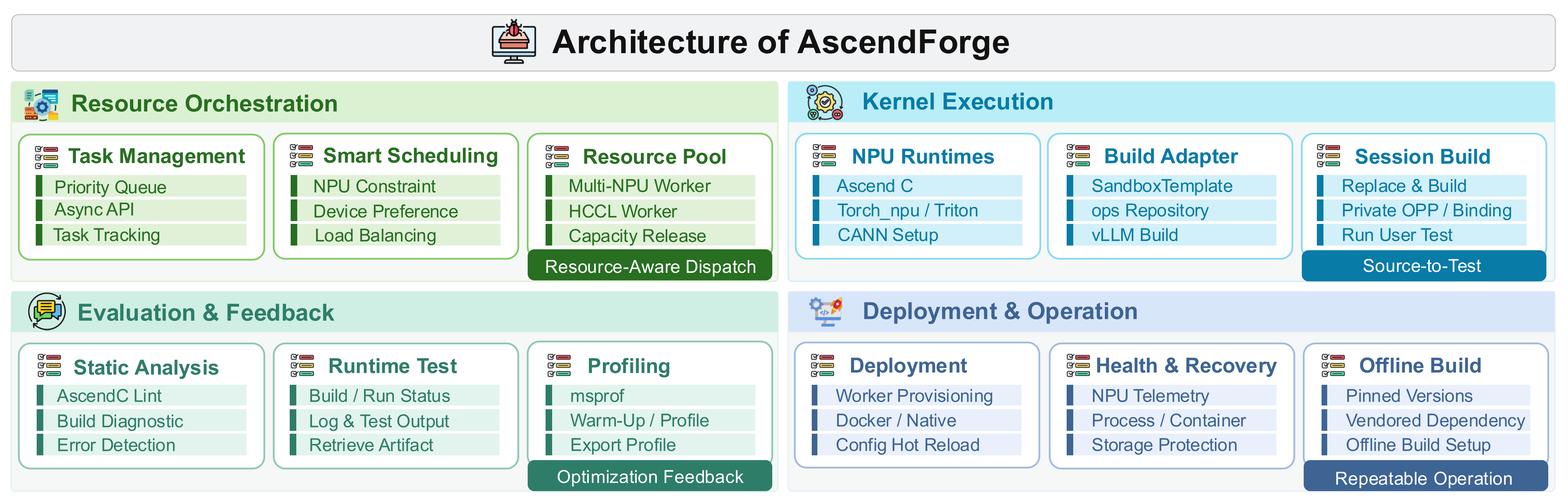}}
  \caption{Architecture of AscendForge, integrating resource orchestration, kernel execution, evaluation and feedback, and deployment management for Ascend kernel optimization.}
  \label{fig:ascendforge}
\end{figure}

AscendGYM includes AscendForge, an Ascend-native execution sandbox for candidate compilation, correctness checking, and performance measurement.

Like other code evaluation benchmarks, AscendGYM must verify that implementations preserve the required computational semantics after successful compilation and execution. KernelBench~\citep{ouyang2025kernelbench} demonstrates the importance of executing candidate implementations and returning compiler feedback and correctness results to the model. SOL-ExecBench~\citep{lin2026solexecbench} incorporates isolated execution and checks against inadvertent modifications to the measurement procedure into its evaluation harness. FullStack Bench~\citep{cheng2024fullstack} released SandboxFusion as reusable infrastructure for executing and evaluating generated programs across programming languages. Building on SandboxFusion~\citep{cheng2024fullstack}, AscendForge introduces adapters that manage dependencies among host-side registration, device-side kernels, the target system-on-chip (SoC), and custom kernel packages, enabling compilation and Python interface binding for NPU devices.

As shown in Figure~\ref{fig:ascendforge}, AscendForge comprises four layers for kernel building and execution, resource scheduling, development feedback, and deployment and operations.

\noindent\textbf{Execution and Build Layer.} This layer supports Ascend C and Python NPU execution backends, with adapters for CANN kernel repositories~\citep{cann_ops_math,cann_ops_cv,cann_ops_nn,cann_ops_transformer}, an Ascend C custom kernel build template, and projects such as vLLM Ascend~\citep{vllm_ascend}. It automates source submission, compilation and installation, Python interface binding, and test execution.

\noindent\textbf{Resource Orchestration Layer.} This layer manages evaluation requests through a scheduling proxy, task queues, and multiple execution nodes. It allocates resources according to the required NPU count, device model, and node preferences, and supports concurrent evaluation through resource reservations and load management.

\noindent\textbf{Development Feedback Layer.} This layer integrates static code analysis and performance profiling, and returns build logs, correctness results, performance measurements, and test artifacts. These outputs help agents diagnose errors, identify performance bottlenecks, and guide the next optimization iteration.

\noindent\textbf{Deployment and Operations Layer.} This layer provides hardware health monitoring, configuration hot reloading, and automatic process recovery, together with offline dependency preparation and version verification, to support service availability and environment consistency.

The interface design of AscendForge preserves the build conventions of the underlying projects and provides a session-specific workspace for each build. AscendForge separates the client, scheduling proxy, and execution workers. The client packages the source code and test driver, and the scheduling proxy selects a compatible worker using device count, SoC type, and worker group as scheduling constraints. Node and device preferences are also considered when this information is available. The worker builds and executes the kernel; the test driver first verifies correctness and then collects performance metrics. Compilation and execution results are reported separately during testing, allowing the model to distinguish build failures, numerical accuracy failures, and performance regressions. The harness uses this feedback to guide the next optimization iteration, while AscendGYM reevaluates the candidate implementation at the end of the current run.

\subsection{Metric Design}
\label{sec:metrics}

While conventional kernel benchmarks primarily evaluate final
performance, AscendGYM evaluates both optimization outcomes and
the resources required to achieve them. Its evaluation framework
includes joint performance--budget metrics, standalone performance
and budget metrics, and a reliability metric. In particular, the
joint metrics capture optimization quality and efficiency within
a single score.

\subsubsection{Joint Performance--Budget Metrics}

\paragraph{Speedup--Token Score (STS).}
AscendGYM evaluates the kernel optimization capabilities of LLMs
and agent harnesses under explicit resource budgets, reflecting
the practical constraints of kernel optimization. STS jointly
measures the best speedup achieved within a token budget and
the token consumption required to obtain improvements:
\begin{equation}
\label{eq:sts}
\mathrm{STS}
=
\underbrace{100\log_2 S_n}_{\text{final performance gain}}
-
\underbrace{
50\sum_{k=1}^{n}
\frac{B_k}{B_{\max}}\Delta g_k
}_{\text{penalty for delayed improvements}}
\end{equation}
Here, $B_{\max}>0$ is the token budget, and $n$ is the number
of candidate evaluations completed within that budget.
$B_k$ denotes the cumulative token consumption when evaluation
$k$ completes, including tokens spent on failed attempts and
attempts that yield no improvement.
$S_k$ denotes the best correctness-validated speedup available
after evaluation $k$, relative to the reference implementation.
The reference implementation is retained when no improvement
is found, so $S_0=1$ and $S_k\ge1$.

The incremental log-speedup gain is
\begin{equation}
\Delta g_k
=
\log_2 S_k-\log_2 S_{k-1}.
\end{equation}
The first term in Eq.~\eqref{eq:sts} rewards final performance,
whereas the second penalizes improvements obtained later in
the token budget. Thus, for the same final speedup, earlier
improvements receive a higher score. Higher STS indicates
better budget-aware optimization under the same evaluation
conditions. If no improvement is achieved, $\mathrm{STS}=0$.

\paragraph{Cost-to-Gain (CTG).}
CTG measures the additional token consumption required to
increase speedup by a specified fraction relative to a reference
checkpoint:
\begin{equation}
\label{eq:ctg}
\mathrm{CTG}(\delta;r)
=
\min_{\substack{r\le k\le n\\
                 S_k\ge(1+\delta)S_r}}
\left(B_k-B_r\right),
\qquad
\delta\in\{0.01,0.05,0.10\}.
\end{equation}
Here, $r$ is the reference checkpoint, and $B_k-B_r$ is the
token consumption accumulated since that checkpoint.
The thresholds correspond to relative speedup gains of
$1\%$, $5\%$, and $10\%$.
We define the minimum of an empty set as $+\infty$,
indicating that the target gain was not reached within
the budget. Lower CTG indicates greater token efficiency.

\subsubsection{Performance Metrics}

AscendGYM reports GM Speedup and the $\mathrm{fast}_p$
metric introduced by KernelBench~\citep{ouyang2025kernelbench}.
For $N$ evaluation tasks, let $S_{i,n_i}$ denote the best
correctness-validated speedup achieved on task $i$ within
the fixed budget.

\paragraph{$\mathbf{fast}_{\mathbf{p}}$.}
$\mathrm{fast}_p$ measures the fraction of tasks for which
a correct implementation achieves a speedup strictly greater
than $p$:
\begin{equation}
\mathrm{fast}_p
=
\frac{1}{N}
\sum_{i=1}^{N}
\mathds{1}\!\left[S_{i,n_i}>p\right],
\qquad p\ge1.
\end{equation}
Here, $\mathds{1}[\cdot]$ is the indicator function.
Correctness is implicit in the definition of $S_{i,n_i}$.
Tasks with no improvement retain a speedup of $1$ and
therefore do not contribute to $\mathrm{fast}_p$ for $p\ge1$.

\paragraph{Geometric Mean Speedup (GM Speedup).}
GM Speedup summarizes performance across tasks using their geometric mean:
\begin{equation}
\mathrm{GM\ Speedup}
=
\left(\prod_{i=1}^{N}S_{i,n_i}\right)^{1/N}.
\end{equation}
Each task receives equal weight, and the reference implementation
is retained when no improvement is found.

\subsubsection{Budget Metrics}

AscendGYM measures optimization cost using cumulative and mean token consumption, LLM interaction rounds, and wall-clock time. All budget metrics include failed attempts to capture the full resource and time costs of obtaining optimization results.

\subsubsection{Reliability Metrics}

\paragraph{Candidate Pass Rate.}
Candidate Pass Rate measures how reliably an LLM or agent harness produces correct candidates. It is defined as the fraction of evaluated candidates that successfully compile, execute, and pass all required correctness checks. Compilation failures, runtime errors, timeouts, and incorrect outputs count as failures.

\paragraph{Task Pass Rate.}
Task Pass Rate measures the fraction of attempted tasks with a valid baseline measurement and at least one candidate that passes the required correctness checks during search. Tasks with no candidate submission or no passing candidate remain in the denominator. This measures the coverage of correct outcomes across tasks, while Candidate Pass Rate measures correctness across individual candidate evaluations.

When independent repeats are available, they can additionally assess the reproducibility and stability of optimization outcomes.

\subsection{Evaluation Workflow}
\label{sec:workflow}
\label{sec:user-workflow}

\subsubsection{Evaluation Modes}

AscendGYM supports two evaluation modes---model--harness evaluation and model evaluation---under the same task configurations and validation environment.

\paragraph{Model--Harness Evaluation.}
Users configure an existing coding agent, such as Claude Code, Codex, or OpenCode, or integrate a custom harness. AscendGYM evaluates the resulting model--harness pair as a complete system.

\paragraph{Model Evaluation.}
Users provide an LLM API, and AscendGYM runs the model
with AuraKernel, hereafter referred to as \texttt{Default}, as the default harness. AuraKernel supports strategy search and experience summarization. Holding the \texttt{Default} configuration fixed provides consistent evaluation conditions across models. The configuration of \texttt{Default} is detailed in Appendix~\ref{app:default-harness}.

\subsubsection{Evaluation Procedure}

\begin{figure}[t]
  \centering
  \makebox[\textwidth][c]{\includegraphics[width=1\textwidth]{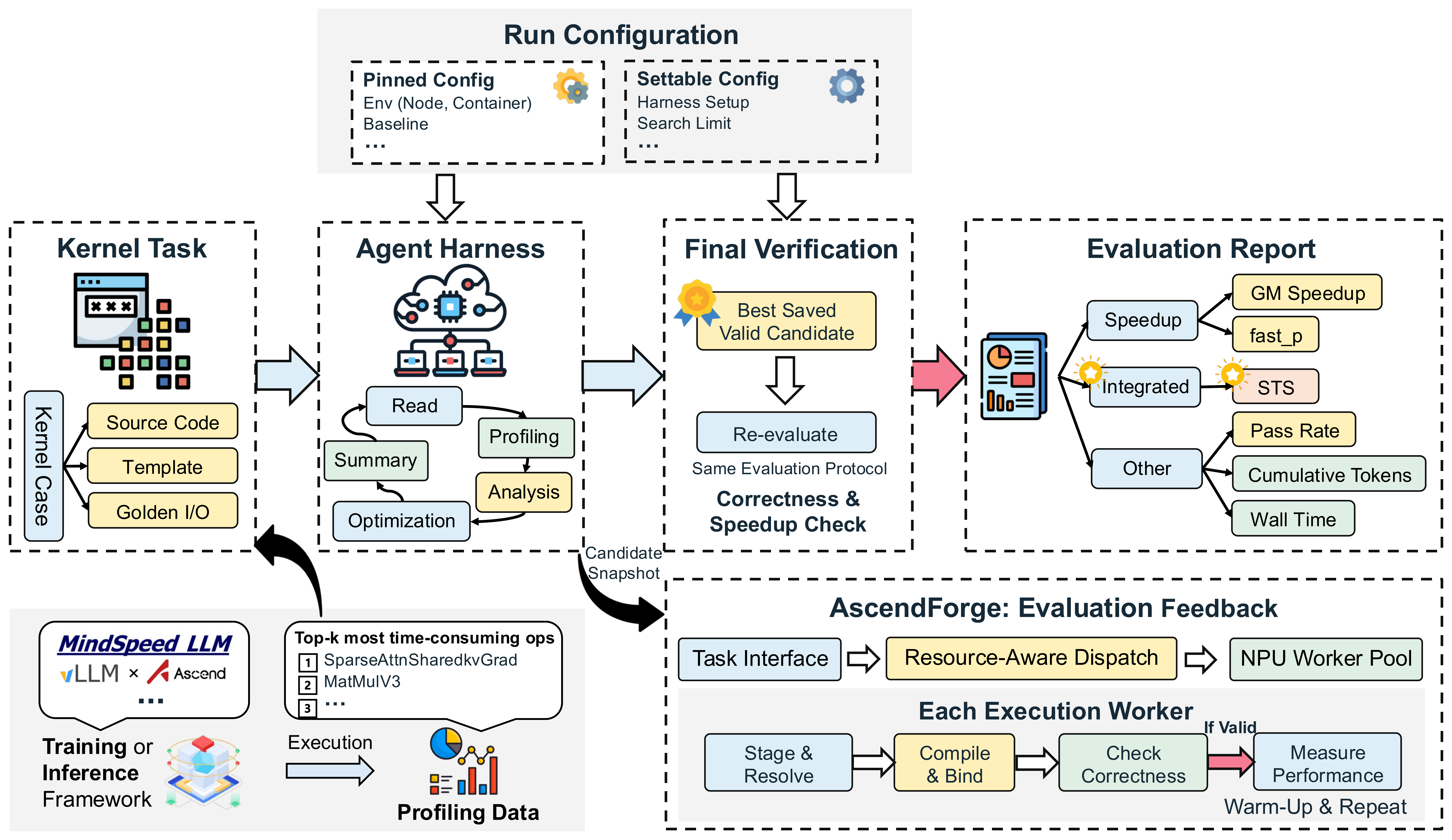}}
  \caption{Overview of the AscendGYM workflow.}
  \label{fig:evaluation-workflow}
\end{figure}

Figure~\ref{fig:evaluation-workflow} illustrates how AscendGYM constructs kernel optimization tasks from real training and inference workloads. We collect profiling data from workloads running in Ascend frameworks, such as MindSpeed-LLM and vLLM-Ascend, and identify the top-$k$ bottleneck kernels. Each kernel optimization task includes the source code of the baseline, template metadata specifying input and output tensor shapes, test inputs, and reference outputs generated by the corresponding CPU implementation. These artifacts support numerical correctness checks and isolated kernel performance measurement.

To launch an evaluation, users configure the execution
environment and optimization settings. AscendGYM first measures baseline performance, then enters an iterative loop of bottleneck analysis, code refinement, performance measurement, and experience summarization. Each optimized candidate is submitted to the AscendForge sandbox for compilation, Python interface binding, and correctness checks. Candidates that pass
these checks are profiled, and the resulting measurements guide subsequent optimization.

Search terminates when the resource budget is exhausted or the maximum iteration count is reached. The system then revalidates the best implementation and uses the performance measurements and optimization traces to compute the metrics defined in Section~\ref{sec:metrics}.

\section{Experiments}
\label{sec:experiments}

\subsection{Setup}
\label{sec:setup}

\paragraph{Evaluated Models and Harnesses.}
We conduct experiments using the two evaluation modes defined in Section~\ref{sec:workflow}.
For model--harness evaluation, we evaluate selected combinations of models and three coding-agent harnesses: Claude Code~\citep{anthropic_claude_code}, Codex CLI~\citep{openai_codex_cli}, and OpenCode~\citep{opencode_docs}.
The evaluated models are Claude Opus 5~\citep{anthropic2026opus5}, GPT-5.6 Terra~\citep{openai2026gpt56terra}, GLM-5.3~\citep{zai2026glm53}, Qwen3.5-122B-A10B~\citep{qwen2026qwen35_122b}, Qwen3.6-35B-A3B~\citep{qwen2026qwen36_35b}, and DeepSeek-V4-Flash~\citep{deepseekai2026deepseekv4}.

For model evaluation, we use the \texttt{Default} harness provided by AscendGYM to evaluate the kernel optimization capabilities of GPT-5.6 Luna~\citep{openai2026gpt56luna} under a fixed workflow and tool interface.
Detailed parameters and context management settings for \texttt{Default} are provided in Appendix~\ref{app:default-harness}.
The shared task prompt for model--harness evaluation is presented in Appendix~\ref{app:evaluation-prompt}.

\paragraph{Tasks and Baselines.}
\label{sec:experiment-scope}

Task sets and test cases adopted in evaluation are described in Section~\ref{sec:task-selection}, with the same initial implementations and test cases across configurations.
For model mode, configuration-specific scoring and measurement protocols are detailed in Appendix~\ref{app:default-harness}.
The evaluated tasks are drawn from CANN \texttt{ops-nn} and \texttt{ops-transformer}~\citep{cann_ops_nn,cann_ops_transformer}, and vLLM Ascend~\citep{vllm_ascend}.
Their task packages specify source versions, test inputs, and numerical comparison rules.

For each task, the baseline is the initial Ascend C implementation shipped with the task package.
We measure its execution latency in the evaluation environment and use this measured latency as the performance reference for candidate implementations.
All speedups are computed relative to the corresponding task's initial implementation.

\paragraph{Optimization Budgets and Final Re-evaluation.}
Each model--harness optimization run allows at most 80 candidate evaluations, with a soft target of exploring 50 substantively different optimization solutions specified in the prompt.
After reaching this target, agents may continue optimizing within the remaining evaluation budget; they may also finish before reaching it if no reasonable optimization direction remains, and explain the reason.
Failed submissions count toward the evaluation budget, whereas baseline measurements are excluded; no hard limits are imposed on total token consumption or wall-clock time.

For these runs, after the search ends, the final selected implementation undergoes five additional performance evaluations.
These evaluations do not consume the search budget, provide no feedback for further optimization, and are not counted as independent optimization runs.

\paragraph{Correctness and Performance Measurement.}
Candidate implementations are validated using each task's predefined test scripts and numerical error criteria.
Only implementations that pass all required checks are considered valid results.

Baseline and candidate implementations follow the same measurement protocol: each evaluation executes eight rounds, with 3 rounds for warm-up and the remaining rounds to compute the arithmetic mean of device-side kernel execution times.
Timings are collected using \texttt{msprof}.
Hardware and software details are provided in Appendix~\ref{app:hardware-software}.

\subsection{Results and Discussion}
\label{sec:results}

Following the workflow in Section~\ref{sec:workflow}, we conduct model--harness evaluation and model evaluation to examine how the metric suite characterizes practical optimization behavior. We focus on how cost constraints affect the assessment of optimization outcomes, how aggregate performance relates to task coverage, and what resource-cost and correctness metrics reveal beyond performance alone. We also use the gains achieved by current systems to assess the challenge of further optimizing existing implementations.

\subsubsection{Model--Harness Evaluation}
\label{sec:results-harness}

Following Section~\ref{sec:metrics}, we analyze the results through four metric categories: joint performance--budget, performance, budget, and reliability metrics.

\begin{figure}[ht]
\centering
{\small\textbf{Tight budget: 10M tokens}\par}
\smallskip
\includegraphics[width=\linewidth]{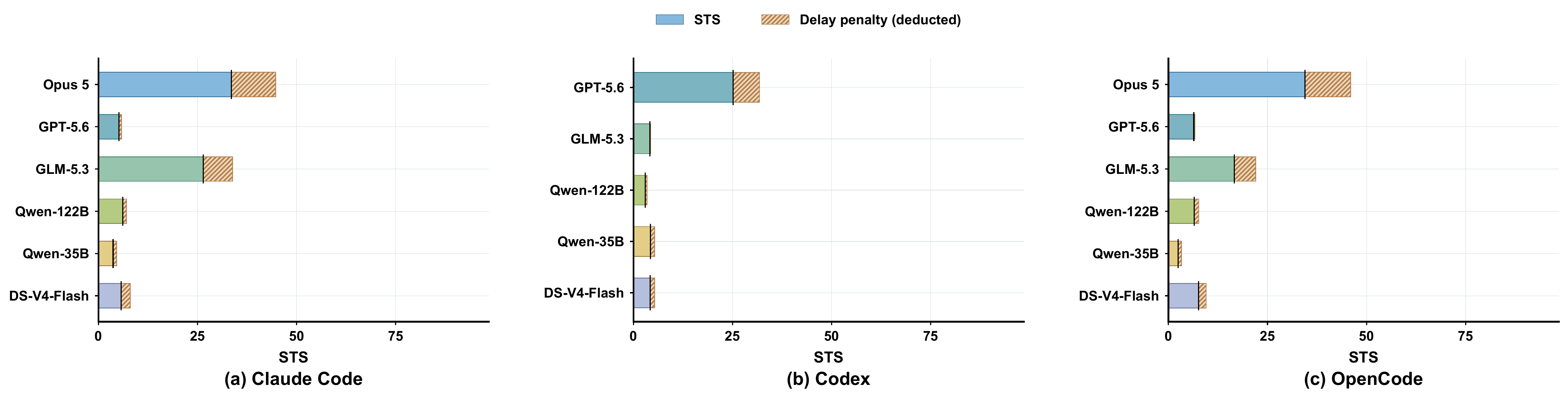}
\par\medskip
{\small\textbf{Moderate budget: 20M tokens}\par}
\smallskip
\includegraphics[width=\linewidth]{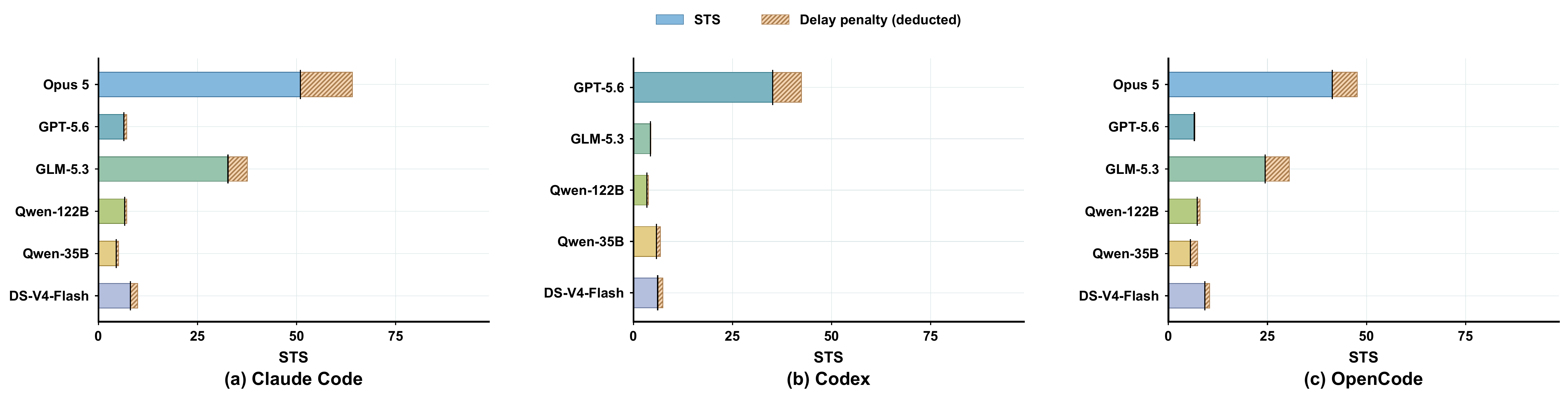}
\par\medskip
{\small\textbf{Generous budget: 50M tokens}\par}
\smallskip
\includegraphics[width=\linewidth]{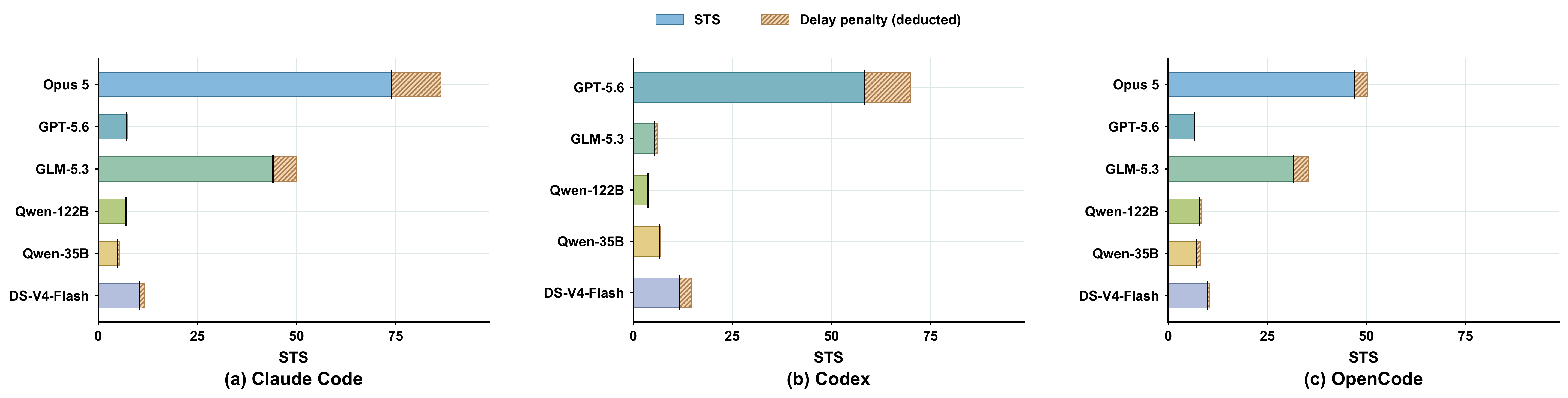}
\caption{STS under three token budgets. Solid bars show scores; hatched segments show cost deductions.}
\label{fig:results-harness-sts-budgets}
\end{figure}

\paragraph{Joint Performance--Budget Metrics.}
\label{sec:results-sts}
Figure~\ref{fig:results-harness-sts-budgets} presents three usage scenarios defined in this study: tight, moderate, and generous per-task budgets of 10M, 20M, and 50M tokens, respectively. Combinations exhibit different gain trajectories as the budget increases. For example, Opus~5 achieves similar STS with Claude Code and OpenCode at 10M tokens. With larger budgets, Claude Code's performance component continues to grow, whereas OpenCode obtains more limited additional gains. Reporting only performance at the end of the full search therefore does not adequately describe which gains are available under different budgets.

Cost deductions further reveal differences in the investment required for these gains. At 50M tokens, Claude Code/Opus~5 has a larger performance component than OpenCode/Opus~5 but also incurs a larger cost deduction. The deduction narrows the score gap without eliminating the former's advantage (Figure~\ref{fig:results-harness-sts-budgets}). Assessing optimization should account for both useful results obtained with limited expenditure and substantial gains enabled by additional investment. Scores and their components across the three budgets make this trade-off visible.

\begin{figure}[ht]
\centering
\includegraphics[width=\linewidth]{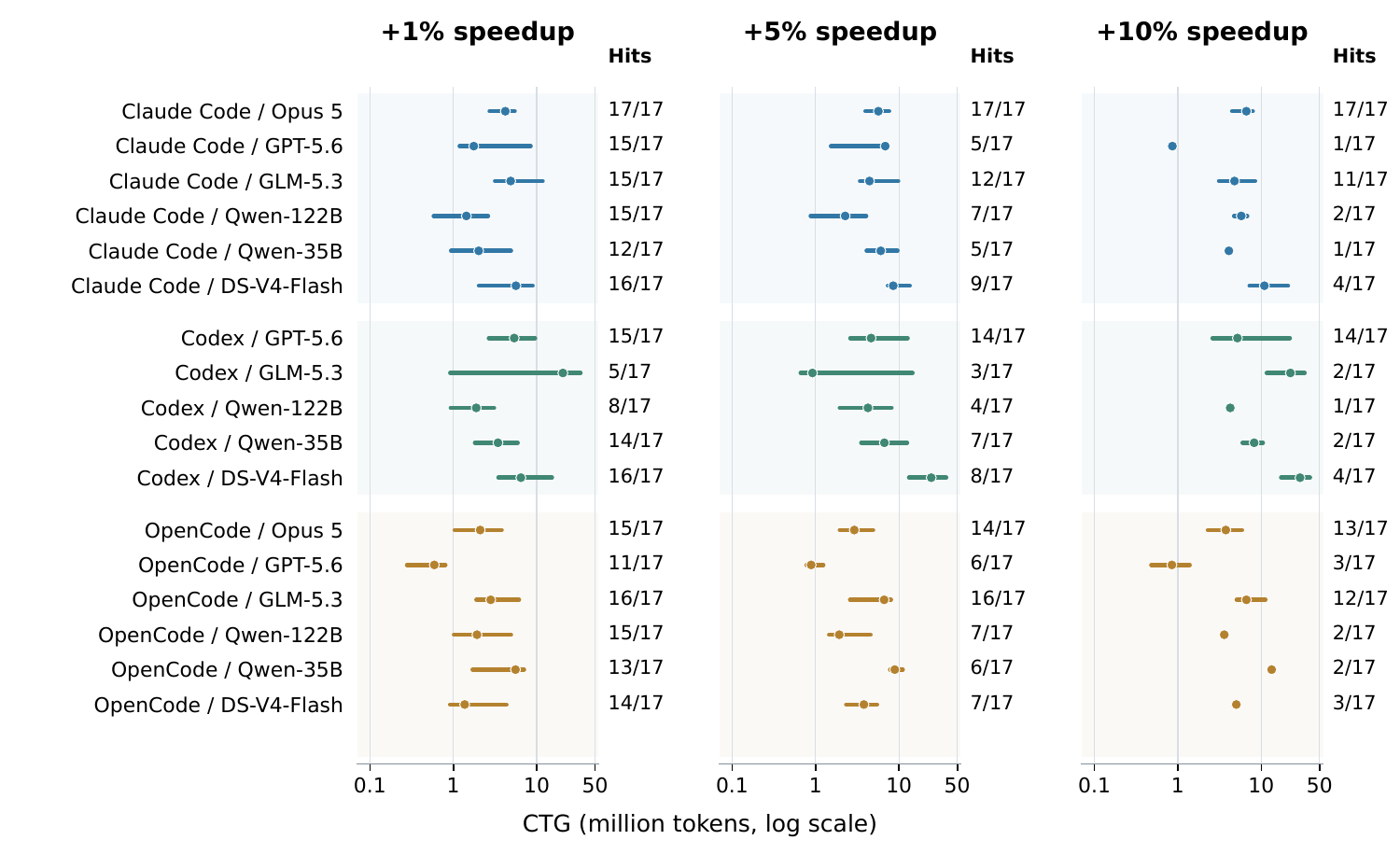}
\caption{CTG at a 50M-token budget. Dots and lines show medians and interquartile ranges over attained tasks; counts give attainment out of 17 tasks in Batch 1.}
\label{fig:results-ctg}
\end{figure}

The CTG results further show that lower attainment cost and broader task coverage do not always coincide. At a 50M-token budget and a $10\%$ gain target, Opus~5 with OpenCode and Claude Code has median costs of 3.8M and 6.6M tokens over attained tasks, respectively, but the former reaches the target on only a subset of tasks while the latter covers all tasks (Figure~\ref{fig:results-ctg}). Because the attained task sets differ, the lower conditional median does not establish greater efficiency on the same tasks. Reporting CTG together with attainment counts captures both the cost and coverage of target gains, avoiding choices based solely on costs among successful tasks.

\begin{figure}[ht]
\centering
\includegraphics[width=\linewidth]{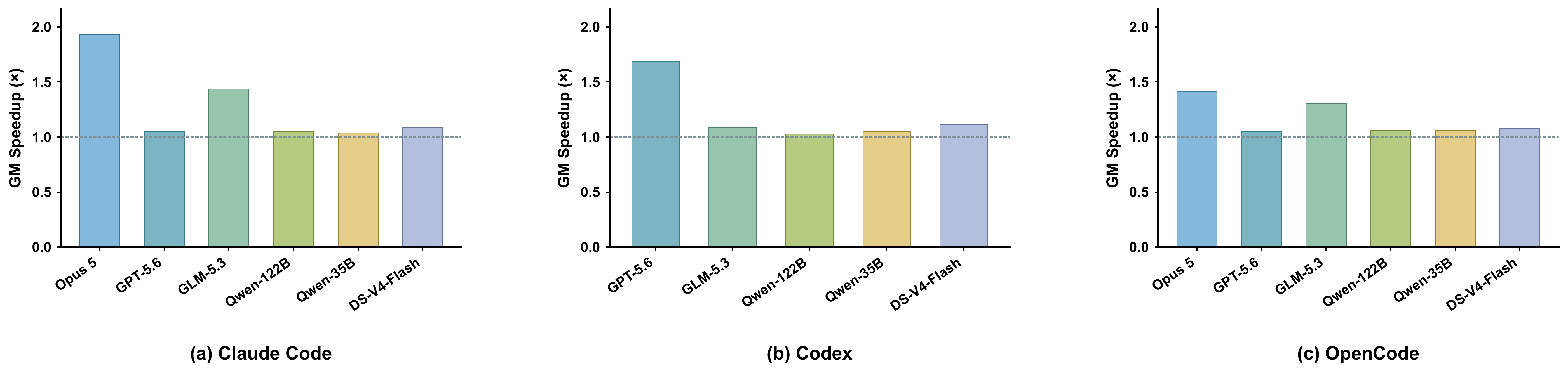}
\caption{GM speedup over full search trajectories, where dashed lines mark $1\times$.}
\label{fig:results-gm-speedup}
\end{figure}

\begin{figure}[ht]
\centering
\includegraphics[width=\linewidth]{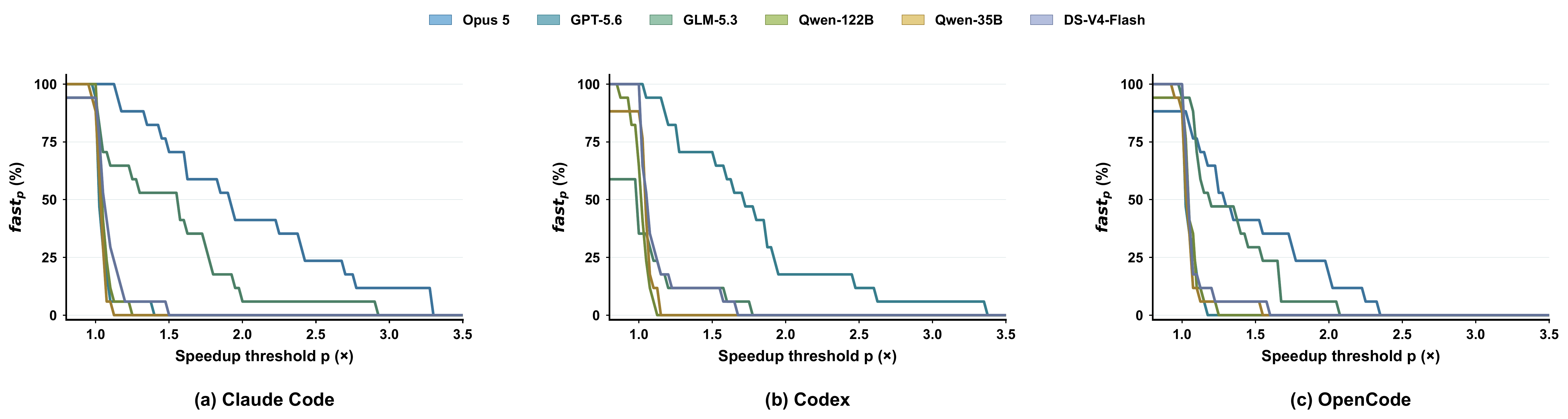}
\caption{Speedup-threshold attainment ($\mathrm{fast}_p$) over full search trajectories.}
\label{fig:results-threshold}
\end{figure}

\paragraph{Performance Metrics.}
\label{sec:results-speedup}
GM speedup and $\mathrm{fast}_p$ jointly reveal differences between aggregate gains and their distribution (Figures~\ref{fig:results-gm-speedup} and~\ref{fig:results-threshold}). Claude Code/Opus~5 and Codex/GPT-5.6 both accelerate all tasks, but seven and three tasks, respectively, exceed $2\times$ speedup. Identical coverage of accelerated tasks therefore does not imply identical improvement magnitudes; increasing the threshold allows $\mathrm{fast}_p$ to distinguish these outcomes. Aggregate speedup should consequently be interpreted alongside coverage at different thresholds to assess whether gains meet application requirements.

Overall, current systems still achieve limited gains on the benchmark. Approximately $70\%$ of the evaluated model--harness combinations have GM speedups below $1.2\times$, with gains largely concentrated in modest improvements and limited task coverage at higher speedup thresholds (Figures~\ref{fig:results-gm-speedup} and~\ref{fig:results-threshold}). Under the given initial implementations and evaluation conditions, achieving substantial improvements across a broad range of tasks remains challenging. These results indicate that AscendGYM poses a meaningful challenge for further optimizing existing kernels. Per-kernel results are provided in Figure~\ref{fig:appendix-harness-per-operator}.

\paragraph{Budget Metrics.}
\label{sec:results-resources}
Token consumption, interaction rounds, and wall-clock time exhibit different patterns across combinations, indicating that a single cost measure cannot fully characterize optimization expenditure. Claude Code/Opus~5 consumes fewer tokens than Codex/GPT-5.6 but uses more interaction rounds. OpenCode/Qwen-35B has relatively modest token and round counts, yet takes approximately 11.1 hours per task (Figures~\ref{fig:appendix-harness-tokens}--\ref{fig:appendix-harness-wall-time}).

Reducing token expenditure and shortening the wait for optimization may therefore favor different choices. Reporting each cost measure separately complements the joint scores and supports decisions under different resource constraints. Detailed statistics are provided in Appendix~\ref{app:harness-resource-costs}; cumulative and mean costs are computed from the selected run records and include non-passing runs.

\begin{figure}[ht]
\centering
\includegraphics[width=\linewidth]{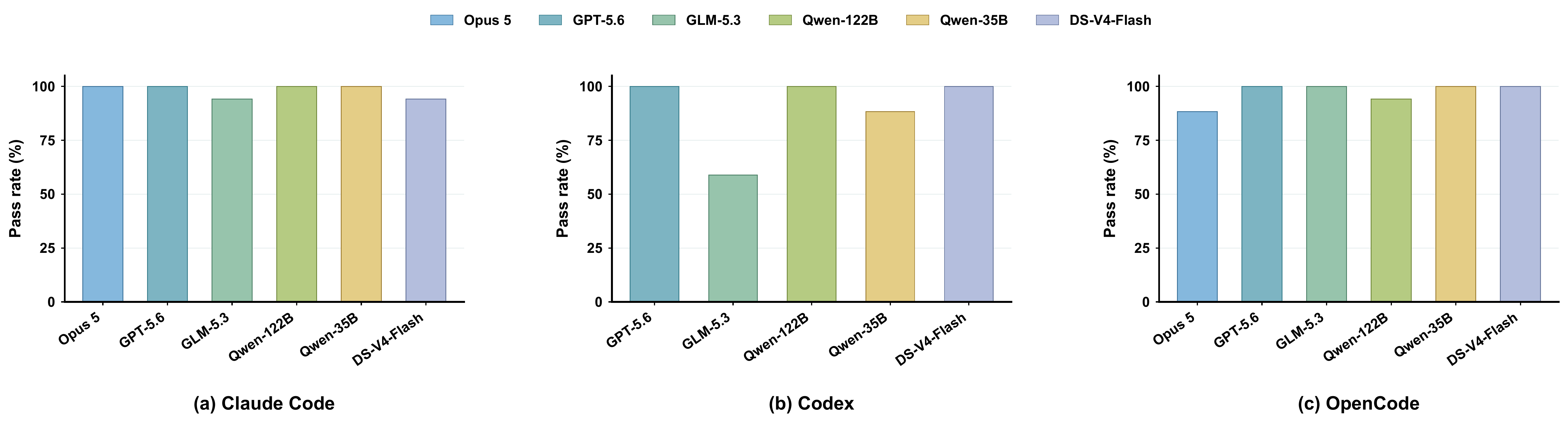}
\caption{The task pass rate among different combinations.}
\label{fig:results-correctness}
\end{figure}

\paragraph{Reliability Metrics.}
\label{sec:results-correctness}
Task pass rate and performance gains provide different information. GPT-5.6 passes all tasks with each of the three harnesses, yet its GM speedup differs substantially: approximately $1.69\times$ with Codex versus $1.05\times$ with Claude Code (Figures~\ref{fig:results-correctness} and~\ref{fig:results-gm-speedup}). Producing correct candidates therefore does not guarantee substantial performance gains, and high pass rates do not imply that the optimization challenges have been resolved.

Some combinations also fail to produce correct candidates on all tasks. Because performance metrics retain the initial implementation when no improvement is found, speedup alone cannot distinguish between a run with no correct candidate and one with correct candidates that do not improve performance. Task pass rate supplies this complementary information, allowing performance results to be interpreted together with the outcomes of the complete evaluation workflow. Task-level accounting is provided in Appendix~\ref{app:harness-reliability}.

\subsubsection{Model Evaluation}
\label{sec:results-model}

We hold the model fixed at GPT-5.6 Luna and compare four \texttt{Default} configurations: cold, warm, no lessons, and code only. The task scope and configuration-specific protocols are specified in Section~\ref{sec:experiment-scope} and Appendix~\ref{app:default-harness}.

\begin{figure}[ht]
\centering
\includegraphics[width=\linewidth]{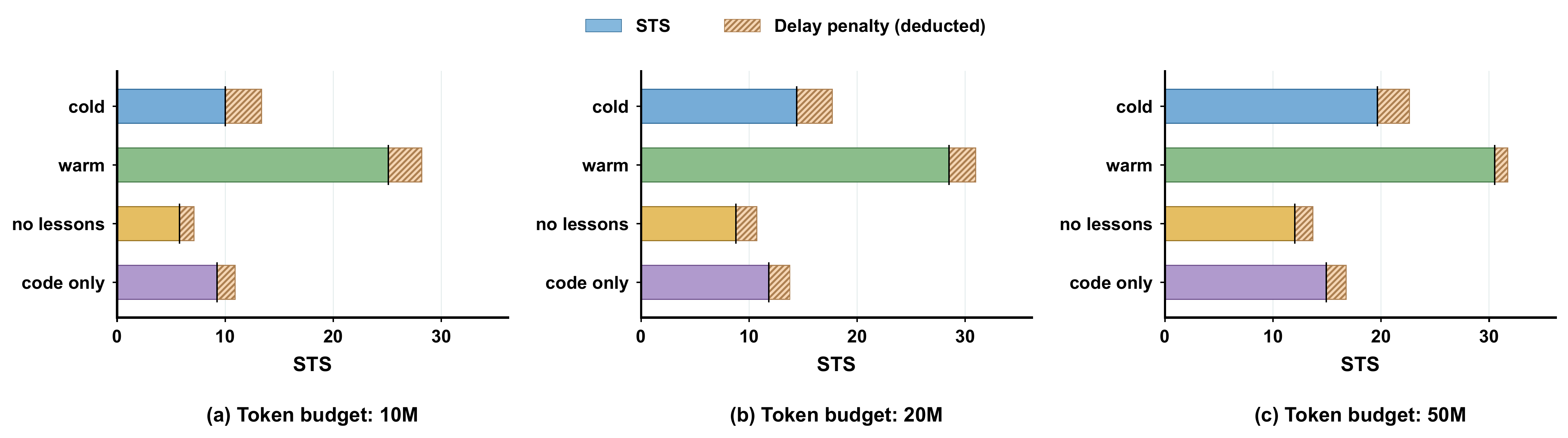}
\caption{STS under three token budgets using GPT-5.6 Luna. Solid bars show scores; hatched segments show cost deductions.}
\label{fig:results-model-sts}
\end{figure}

\paragraph{Joint Performance--Budget Metrics.}
Warm achieves the highest STS at all three budgets (Figure~\ref{fig:results-model-sts}). At 10M tokens, its performance component already exceeds cold's while its cost deduction is smaller, indicating that this configuration obtains greater gains within the limited budget with a lower cost of attaining those improvements.

As the budget increases, cold's performance component grows further and the gap to warm narrows; however, warm retains both a higher score and a larger performance component at 50M tokens (Figure~\ref{fig:results-model-sts}). Under their respective evaluation protocols, continued search with cold recovers part of the gap, whereas warm already provides strong optimization results under a tight budget.

Code only outperforms no lessons at all three budgets, but both remain below the full cold configuration (Figure~\ref{fig:results-model-sts}). These differences persist with the model held fixed, showing that the \texttt{Default} configuration also influences optimization outcomes. The comparison describes each configuration's overall performance under its corresponding evaluation protocol (Appendix~\ref{app:default-harness}). Comparisons between models using \texttt{Default} should therefore keep the settings and prior experience consistent to avoid attributing configuration differences to model capability.

\section{Conclusion}
\label{sec:conclusion}

We introduced \textbf{AscendGYM}, a kernel optimization benchmark for Ascend NPUs that evaluates agents' ability to further improve existing implementations.
Its initial suite comprises 48 kernels from real LLM workloads across three difficulty levels, each defining an optimization task with complete, runnable initial source code and input configurations extracted from the corresponding workloads.
Through its built-in execution sandbox, \textbf{AscendForge}, AscendGYM provides consistent candidate compilation, correctness checking, and performance measurement, supporting both model comparisons using a fixed Default harness and evaluation of model--harness combinations.
We also designed \textbf{STS}, which combines final performance gains with cumulative token consumption at each improvement to jointly assess optimization effectiveness and process efficiency.

We systematically evaluated multiple leading models paired with different harnesses on the 17 tasks in Batch~1 of this suite.
The results support the benchmark and metric design by showing that the tasks remain challenging for current agents and that jointly assessing performance, cost, and reliability reveals distinctions that speedup alone does not capture.
Future work will extend the evaluation to the remaining 31 tasks in Batches~2 and~3 and broaden the task and input coverage to further assess agents' kernel optimization capabilities in real LLM training and inference scenarios.

% Uncomment these lines after adding citations to references.bib.
\bibliographystyle{slai}
\bibliography{references}

\begin{thebibliography}{36}
\providecommand{\natexlab}[1]{#1}
\providecommand{\url}[1]{\texttt{#1}}
\expandafter\ifx\csname urlstyle\endcsname\relax
  \providecommand{\doi}[1]{doi: #1}\else
  \providecommand{\doi}{doi: \begingroup \urlstyle{rm}\Url}\fi

\bibitem[Andrews \& Witteveen(2025)Andrews and Witteveen]{andrews2025gpu}
Martin Andrews and Sam Witteveen.
\newblock Gpu kernel scientist: An llm-driven framework for iterative kernel
  optimization.
\newblock \emph{arXiv preprint arXiv:2506.20807}, 2025.

\bibitem[{Anthropic}(2026)]{anthropic2026opus5}
{Anthropic}.
\newblock Introducing {Claude Opus 5}, 2026.
\newblock URL \url{https://www.anthropic.com/news/claude-opus-5}.
\newblock Accessed: 2026-10-19.

\bibitem[{Anthropic}(n.d.)]{anthropic_claude_code}
{Anthropic}.
\newblock {Claude Code}: Overview.
\newblock Official documentation, n.d.
\newblock URL \url{https://code.claude.com/docs/en/overview}.
\newblock Accessed: 2026-10-19.

\bibitem[Baronio et~al.(2026)Baronio, Marsella, Pan, Guo, and
  Alberti]{baronio2026kevin}
Carlo Baronio, Pietro Marsella, Ben Pan, Simon Guo, and Silas Alberti.
\newblock Kevin: Multi-turn rl for generating cuda kernels.
\newblock In \emph{International Conference on Learning Representations},
  volume 2026, pp.\  83418--83452, 2026.

\bibitem[{CANN\_ops\_cv}(n.d.)]{cann_ops_cv}
{CANN\_ops\_cv}.
\newblock {ops-cv}: {CANN} computer vision operator library.
\newblock GitCode repository, n.d.
\newblock URL \url{https://gitcode.com/cann/ops-cv}.
\newblock Accessed: 2026-10-11.

\bibitem[{CANN\_ops\_math}(n.d.)]{cann_ops_math}
{CANN\_ops\_math}.
\newblock {ops-math}: {CANN} math operator library.
\newblock GitCode repository, n.d.
\newblock URL \url{https://gitcode.com/cann/ops-math}.
\newblock Accessed: 2026-10-11.

\bibitem[{CANN\_ops\_nn}(n.d.)]{cann_ops_nn}
{CANN\_ops\_nn}.
\newblock {ops-nn}: {CANN} neural network operator library.
\newblock GitCode repository, n.d.
\newblock URL \url{https://gitcode.com/cann/ops-nn}.
\newblock Accessed: 2026-10-11.

\bibitem[{CANN\_ops\_transformer}(n.d.)]{cann_ops_transformer}
{CANN\_ops\_transformer}.
\newblock {ops-transformer}: {CANN} transformer operator library.
\newblock GitCode repository, n.d.
\newblock URL \url{https://gitcode.com/cann/ops-transformer}.
\newblock Accessed: 2026-10-11.

\bibitem[Cao et~al.(2026)Cao, Zhai, Li, Hu, Yan, Mu, Fang, She, Li, and Su
  et~al.]{cao2026ascendkernelgen}
Xinzi Cao, Jianyang Zhai, Pengfei Li, Zhiheng Hu, Cen Yan, Bingxu Mu, Guanghuan
  Fang, Bin She, Jiayu Li, Yihan Su, Dongyang Tao, Xiansong Huang, Fan Xu,
  Feidiao Yang, Yao Lu, Chang-Dong Wang, Yutong Lu, Weicheng Xue, Bin Zhou, and
  Yonghong Tian.
\newblock {AscendKernelGen}: A systematic study of {LLM}-based kernel
  generation for neural processing units, 2026.
\newblock URL \url{https://arxiv.org/abs/2601.07160}.

\bibitem[Cha et~al.(2025)Cha, Lee, Kwon, Lee, and Kwon]{cha2025multi}
JooHyoung Cha, Munyoung Lee, Jinse Kwon, Jemin Lee, and Yongin Kwon.
\newblock Multi-level machine learning-guided autotuning for efficient code
  generation on a deep learning accelerator.
\newblock In \emph{Proceedings of the 26th ACM SIGPLAN/SIGBED International
  Conference on Languages, Compilers, and Tools for Embedded Systems}, pp.\
  134--145, 2025.

\bibitem[Cheng et~al.(2024)Cheng, Chen, Chen, Chen, Chen, Chen, Chen, Geng, Li,
  and Li et~al.]{cheng2024fullstack}
Yao Cheng, Jianfeng Chen, Jie Chen, Li~Chen, Liyu Chen, Wentao Chen, Zhengyu
  Chen, Shijie Geng, Aoyan Li, Bo~Li, et~al.
\newblock {FullStack Bench}: Evaluating {LLMs} as full stack coders.
\newblock \emph{arXiv preprint arXiv:2412.00535}, 2024.

\bibitem[{DeepSeek-AI} et~al.(2026)]{deepseekai2026deepseekv4}
{DeepSeek-AI} et~al.
\newblock {DeepSeek-V4}: Towards highly efficient million-token context
  intelligence, 2026.
\newblock URL \url{https://arxiv.org/abs/2606.19348}.

\bibitem[Gao et~al.(2026)Gao, Pan, Su, Li, Du, Zhu, Ma, Fan, Liao, and Hu
  et~al.]{gao2026cannbench}
Xue-Jian Gao, Deng Pan, Yueming Su, Jiasheng Li, Bin Du, Fengming Zhu, Chengdi
  Ma, Junyi Fan, Qichen Liao, Chengqiu Hu, Xinxian Chen, Lingchao Zheng, Jun
  Li, Jiwei Yang, and Yuwei Fan.
\newblock {CANN Bench}: Benchmarking agent generated kernels against real {NPU}
  and algorithmic limits, 2026.
\newblock URL \url{https://arxiv.org/abs/2607.20518}.

\bibitem[Jaber \& Jaber(2026)Jaber and Jaber]{jaber2026autokernel}
Jaber Jaber and Osama Jaber.
\newblock Autokernel: Autonomous gpu kernel optimization via iterative
  agent-driven search.
\newblock \emph{arXiv preprint arXiv:2603.21331}, 2026.

\bibitem[Kamath et~al.(2025)Kamath, Prabhu, Mohan, Peter, Ramjee, and
  Panwar]{kamath2025pod}
Aditya~K Kamath, Ramya Prabhu, Jayashree Mohan, Simon Peter, Ramachandran
  Ramjee, and Ashish Panwar.
\newblock Pod-attention: Unlocking full prefill-decode overlap for faster llm
  inference.
\newblock In \emph{Proceedings of the 30th ACM International Conference on
  Architectural Support for Programming Languages and Operating Systems, Volume
  2}, pp.\  897--912, 2025.

\bibitem[Lei et~al.(2025)Lei, Yang, Zhang, You, Zhang, Luan, Liu, and
  Qian]{lei2025pragma}
Kelun Lei, Hailong Yang, Huaitao Zhang, Xin You, Kaige Zhang, Zhongzhi Luan,
  Yi~Liu, and Depei Qian.
\newblock {PRAGMA}: A profiling-reasoned multi-agent framework for automatic
  kernel optimization, 2025.
\newblock URL \url{https://arxiv.org/abs/2511.06345}.

\bibitem[Li et~al.(2026)Li, Luo, Sun, et~al.]{li2026trex}
Dongfang Li, Xiaodong Luo, Ruoyu Sun, et~al.
\newblock {SLAI T-Rex}: Full-parameter post-training of the {DeepSeek-V4}
  family on {Ascend SuperPOD}, 2026.
\newblock URL \url{https://arxiv.org/abs/2607.20145}.

\bibitem[Li et~al.(2025{\natexlab{a}})Li, Li, Gao, Shi, Li, Wang, Huang, Wang,
  Wang, and Han et~al.]{li2025tritonbench}
Jianling Li, Shangzhan Li, Zhenye Gao, Qi~Shi, Yuxuan Li, Zefan Wang, Jiacheng
  Huang, Haojie Wang, Jianrong Wang, Xu~Han, Zhiyuan Liu, and Maosong Sun.
\newblock {TritonBench}: Benchmarking large language model capabilities for
  generating {Triton} operators.
\newblock In \emph{Findings of the Association for Computational Linguistics:
  ACL 2025}, pp.\  23053--23066. Association for Computational Linguistics,
  2025{\natexlab{a}}.
\newblock \doi{10.18653/v1/2025.findings-acl.1183}.
\newblock URL \url{https://aclanthology.org/2025.findings-acl.1183/}.

\bibitem[Li et~al.(2025{\natexlab{b}})Li, Wang, He, Li, Shi, Li, Hu, Che, Han,
  and Liu et~al.]{li2025autotriton}
Shangzhan Li, Zefan Wang, Ye~He, Yuxuan Li, Qi~Shi, Jianling Li, Yonggang Hu,
  Wanxiang Che, Xu~Han, Zhiyuan Liu, et~al.
\newblock Autotriton: Automatic triton programming with reinforcement learning
  in llms.
\newblock \emph{arXiv preprint arXiv:2507.05687}, 2025{\natexlab{b}}.

\bibitem[Lin et~al.(2026)Lin, Modi, Hari, Huang, Ye, Qin, Zhou, Zhang, Wang,
  and Damani et~al.]{lin2026solexecbench}
Edward Lin, Sahil Modi, Siva Kumar~Sastry Hari, Qijing Huang, Zhifan Ye, Nestor
  Qin, Fengzhe Zhou, Yuan Zhang, Jingquan Wang, Sana Damani, et~al.
\newblock {SOL-ExecBench}: Speed-of-light benchmarking for real-world {GPU}
  kernels against hardware limits, 2026.
\newblock URL \url{https://arxiv.org/abs/2603.19173}.

\bibitem[{OpenAI}(2026{\natexlab{a}})]{openai2026gpt56luna}
{OpenAI}.
\newblock {GPT-5.6 Luna} model.
\newblock Official model documentation, 2026{\natexlab{a}}.
\newblock URL \url{https://developers.openai.com/api/docs/models/gpt-5.6-luna}.
\newblock Accessed: 2026-10-19.

\bibitem[{OpenAI}(2026{\natexlab{b}})]{openai2026gpt56terra}
{OpenAI}.
\newblock {GPT-5.6 Terra} model.
\newblock Official model documentation, 2026{\natexlab{b}}.
\newblock URL
  \url{https://developers.openai.com/api/docs/models/gpt-5.6-terra}.
\newblock Accessed: 2026-10-19.

\bibitem[{OpenAI}(n.d.)]{openai_codex_cli}
{OpenAI}.
\newblock {Codex CLI}.
\newblock Official documentation, n.d.
\newblock URL \url{https://developers.openai.com/codex/cli/}.
\newblock Accessed: 2026-10-19.

\bibitem[{OpenCode Contributors}(n.d.)]{opencode_docs}
{OpenCode Contributors}.
\newblock {OpenCode}: Intro.
\newblock Official documentation, n.d.
\newblock URL \url{https://opencode.ai/docs/}.
\newblock Accessed: 2026-10-19.

\bibitem[Ouyang et~al.(2025)Ouyang, Guo, Arora, Zhang, Hu, R{\'e}, and
  Mirhoseini]{ouyang2025kernelbench}
Anne Ouyang, Simon Guo, Simran Arora, Alex~L Zhang, William Hu, Christopher
  R{\'e}, and Azalia Mirhoseini.
\newblock {KernelBench}: Can {LLMs} write efficient {GPU} kernels?
\newblock In \emph{Proceedings of the 42nd International Conference on Machine
  Learning}, volume 267 of \emph{Proceedings of Machine Learning Research},
  pp.\  47356--47415. PMLR, 2025.
\newblock URL \url{https://proceedings.mlr.press/v267/ouyang25a.html}.

\bibitem[{Qwen Team}(2026{\natexlab{a}})]{qwen2026qwen35_122b}
{Qwen Team}.
\newblock {Qwen3.5-122B-A10B}.
\newblock Hugging Face model card, 2026{\natexlab{a}}.
\newblock URL \url{https://huggingface.co/Qwen/Qwen3.5-122B-A10B}.
\newblock Accessed: 2026-10-19.

\bibitem[{Qwen Team}(2026{\natexlab{b}})]{qwen2026qwen36_35b}
{Qwen Team}.
\newblock {Qwen3.6-35B-A3B}.
\newblock Hugging Face model card, 2026{\natexlab{b}}.
\newblock URL \url{https://huggingface.co/Qwen/Qwen3.6-35B-A3B}.
\newblock Accessed: 2026-10-19.

\bibitem[Tschand et~al.(2025)Tschand, Awad, Swann, Ramakrishnan, Ma, Lowery,
  Dasika, and Reddi]{tschand2025swizzleperf}
Arya Tschand, Muhammad Awad, Ryan Swann, Kesavan Ramakrishnan, Jeffrey Ma,
  Keith Lowery, Ganesh Dasika, and Vijay~Janapa Reddi.
\newblock Swizzleperf: Hardware-aware llms for gpu kernel performance
  optimization.
\newblock \emph{arXiv preprint arXiv:2508.20258}, 2025.

\bibitem[{vLLM Ascend Contributors}(n.d.)]{vllm_ascend}
{vLLM Ascend Contributors}.
\newblock {vLLM Ascend}: A hardware plugin for {vLLM} on {Huawei Ascend}.
\newblock GitHub repository, n.d.
\newblock URL \url{https://github.com/vllm-project/vllm-ascend}.
\newblock Accessed: 2026-10-11.

\bibitem[Wang et~al.(2025)Wang, Majumder, Pandey, Xu, Ding, Rao, Liu, Brahma,
  Li, and Liu et~al.]{wang2025geaktritonv2}
Jianghui Wang, Saptarshi Majumder, Umang Pandey, Chao Xu, Bin Ding, Manoj Rao,
  Ziqiong Liu, Pratik~Prabhanjan Brahma, Dong Li, Zicheng Liu, and Emad
  Barsoum.
\newblock {GEAK-Triton v2} family of {AI} agents: Kernel optimization for {AMD
  Instinct GPUs}.
\newblock AMD ROCm Blog, December 2025.
\newblock URL
  \url{https://rocm.blogs.amd.com/artificial-intelligence/geak-agents-family/README.html}.
\newblock Accessed: 2026-10-20.

\bibitem[Wen et~al.(2025)Wen, Zhang, Li, Liu, Xie, and
  Zhang]{wen2025multikernelbench}
Zhongzhen Wen, Yinghui Zhang, Zhong Li, Zhongxin Liu, Linna Xie, and Tian
  Zhang.
\newblock {MultiKernelBench}: A multi-platform benchmark for kernel generation,
  2025.
\newblock URL \url{https://arxiv.org/abs/2507.17773}.

\bibitem[Xing et~al.(2026)Xing, Zhai, Jiang, Dong, Wu, Ye, Ruan, Huang, Zhang,
  and Yin et~al.]{xing2026flashinferbench}
Shanli Xing, Yiyan Zhai, Alexander Jiang, Yixin Dong, Yong Wu, Zihao Ye,
  Charlie Ruan, Yingyi Huang, Yineng Zhang, Liangsheng Yin, Aksara Bayyapu,
  Luis Ceze, and Tianqi Chen.
\newblock {FlashInfer-Bench}: Building the virtuous cycle for {AI}-driven {LLM}
  systems, 2026.
\newblock URL \url{https://arxiv.org/abs/2601.00227}.

\bibitem[Yang et~al.(2025)Yang, Li, Yang, Zhang, Hui, Zheng, Yu, Gao, Huang,
  and Lv et~al.]{yang2025qwen3}
An~Yang, Anfeng Li, Baosong Yang, Beichen Zhang, Binyuan Hui, Bo~Zheng, Bowen
  Yu, Chang Gao, Chengen Huang, Chenxu Lv, et~al.
\newblock Qwen3 technical report.
\newblock \emph{arXiv preprint arXiv:2505.09388}, 2025.

\bibitem[Yang et~al.(2026)Yang, Wang, Liang, Yang, Ying, You, Zhang, Wang, Yu,
  and Yang et~al.]{yang2026atrexbench}
Lingyun Yang, Yuxiao Wang, Shenghao Liang, Linfeng Yang, Daocheng Ying, Chunbo
  You, Rui Zhang, Luping Wang, Yinghao Yu, Guodong Yang, and Liping Zhang.
\newblock Are {LLM}-generated {GPU} kernels production-ready? a trace-driven
  benchmark and optimization agent, 2026.
\newblock URL \url{https://arxiv.org/abs/2607.14541}.

\bibitem[Ye et~al.(2025)Ye, Chen, Lai, Lin, Zhang, Wang, Chen, Kasikci, Grover,
  and Krishnamurthy et~al.]{ye2025flashinfer}
Zihao Ye, Lequn Chen, Ruihang Lai, Wuwei Lin, Yineng Zhang, Stephanie Wang,
  Tianqi Chen, Baris Kasikci, Vinod Grover, Arvind Krishnamurthy, et~al.
\newblock Flashinfer: Efficient and customizable attention engine for llm
  inference serving.
\newblock \emph{Proceedings of Machine Learning and Systems}, 7, 2025.

\bibitem[{Z.ai}(2026)]{zai2026glm53}
{Z.ai}.
\newblock {GLM-5.3}: Frontier coding with emergent cyber capabilities, 2026.
\newblock URL \url{https://z.ai/blog/glm-5.3}.
\newblock Accessed: 2026-10-19.

\end{thebibliography}

\appendix
% Flush main-text floats before the appendix.
\clearpage
\section{Author List}
\paragraph{Core Contributors} Bingwei Zhang, Jing Li

\paragraph{Contributors (Listed alphabetically by surname)} Amir Abdullin, Sihang Chen, Xin Chen, Liangjun Feng, Yingmeng Gao, Tao Guo, Junjie Huang, Dongfang Li, Zhanglian Lin, Jiajun Liu, Zekun Liu, Zhengxuan Lu, Xiaodong Luo, Jian Meng, Ivan Nosov, Linyuan Qiu, Weijian Sun, Wanrui Yan, Zhicheng Zhou

\paragraph{Project Leaders} Baotian Hu, Min Zhang, Haizhou Li, Zhiquan Luo

\section{Default Harness Configuration}
\label{app:default-harness}

\texttt{Default} is based on AuraKernel and uses GPT-5.6 Luna with reasoning effort set to \texttt{xhigh}. The evaluation protocol covers the 17-task suite described in Section~\ref{sec:task-selection} and includes four configurations: cold, warm, no lessons, and code only.

\paragraph{Configurations and Experience Management.}
Table~\ref{tab:default-configurations} summarizes experience initialization, online experience accumulation, and search policies. Agent-search disables dynamic search-graph exploration and joint proposals from multiple parent implementations. Context is reconstructed for each implementation cycle, while search history and enabled experience stores persist across cycles.

\begin{table}[h]
\centering
\small
\setlength{\tabcolsep}{3pt}
\caption{Default configurations.}
\label{tab:default-configurations}
\begin{tabular}{@{}p{0.18\linewidth}p{0.20\linewidth}p{0.30\linewidth}p{0.24\linewidth}@{}}
\toprule
Variant & Initial experience & Online accumulation & Search policy \\
\midrule
Cold & None & Strategy + coding & Full Default \\
Warm & 5 strategy lessons & Strategy + coding & Full Default \\
No lessons & None & None & Agent-search \\
Code only & None & Code only & Agent-search \\
\bottomrule
\end{tabular}
\end{table}

For CANN tasks, warm imports five strategy lessons from other kernels. For \path{fused_gdn_gating}, \path{hc_post}, and \path{hc_pre_sinkhorn}, warm imports five validated strategy lessons with positive measured gains from a cold run of the same kernel. Imported lessons are fixed before search. When reusing experience for the same kernel, warm starts from the task's initial implementation without inheriting cold's optimized code or search checkpoint; experience preparation costs and warm search costs are recorded separately.

\paragraph{Execution and Measurement Settings.}
Each implementation cycle allows at most ten profiling calls, with a stagnation window of five and a relative-improvement threshold of $0.005$. The harness does not override temperature or top-$p$ and uses the model service's effective settings.

For CANN tasks, each measurement uses eleven executions in warm and eight in the other configurations. All configurations for the three vLLM Ascend tasks use eight executions. The first three executions serve as warm-up, and the remaining executions are averaged to obtain latency. Configuration comparisons reflect their respective budgets, experience sources, and measurement settings.

\section{Evaluation Prompt}
\label{app:evaluation-prompt}

Model--harness evaluation supplies a shared task prompt to the native coding agent, while retaining each harness's own system prompt and tool interfaces. The following abridged English translation of the original Chinese instructions presents the task context, optimization workflow, and evaluation constraints. Task-specific values and deployment paths are represented by placeholders; operator specifications, source files, and tests are provided separately.

% Sources: native_bench/prompt.txt in the archived harness template, and
% the saved native Codex run prompt clauses in run-prompts.json (2026-10-19).
% This is a translated excerpt, not a replacement for the runtime prompt.
\begin{lstlisting}[basicstyle=\ttfamily\footnotesize,breaklines=true,breakatwhitespace=true,columns=fullflexible,keepspaces=true,showstringspaces=false,frame=single,framesep=5pt]
You are a native coding agent participating in a kernel optimization experiment. Use your own file-reading, search, editing, and terminal tools. Do not output custom single-step tool JSON or wait for an external program to direct each step.

Task: <kernel_name>
Working directory: <workspace>
Operator specification: <spec_path>
Input manifest: <manifest_path>
Editable files: <editable_paths>
Test scripts: <test_paths>

Only the listed kernel source files may be modified. Other manifests, tests, specifications, golden references, and evaluation programs must remain unchanged. Only files in the input manifest enter the build; temporary files are not submitted as candidates.

Use your native terminal tool to run:
  baseline: <baseline_command>
  candidate: <candidate_command>
  status: <status_command>
These commands do not invoke a language model.

Optimization workflow:
1. Read the specification, implementation, and tests to understand the semantics and constraints. Run the baseline before editing the kernel implementation.
2. After the baseline passes, use actual evaluation reports to propose optimizations and edit the kernel source files in the workspace.
3. Finish editing before submitting a candidate. Do not modify source files during evaluation; the evaluator freezes a submission snapshot.
4. For compilation or correctness failures, inspect the diagnostics, fix the problem, and resubmit. Failed evaluations also consume the candidate budget.
5. Every passing version is backed up, even if it is slower. While budget remains and no stop is required, you may continue exploring or revert to an earlier version.
6. When the remaining budget is zero or stop_required=true, stop evaluating, editing, and reverting code. Summarize and finish; the controller delivers the best validated saved snapshot.
7. An interrupted HTTP connection does not imply that evaluation did not run. Retain the request_id and query or retry the original call using the same ID.
8. Do not change tolerances, golden references, test coverage, operator interfaces, shapes, or data types. Never present unvalidated code as the best validated implementation.

Search instructions:
- The candidate evaluation limit is 80; baseline measurements do not consume it.
- Within this limit, aim to implement and evaluate 50 substantively different optimization proposals.
- A proposal is a code change submitted for evaluation, not a written suggestion. Proposals may build on the best validated version; they need not be independent, correct, or faster.
- Compilation and correctness repairs to the same proposal do not count as new proposals, although each evaluation consumes budget. Duplicate code and cosmetic changes do not count as new proposals.
- Continue while budget and reasonable optimization directions remain. Reaching the soft target does not require stopping; hard stopping rules take precedence. Explain the reason for ending early.

Measurement and reporting:
- Follow the fixed profiling protocol: eight executions, including three warm-up executions; use the remaining executions for the arithmetic mean. All executions must pass correctness checks.
- There is no hard token or total wall-clock limit. Token consumption and elapsed time are recorded externally; do not estimate or fabricate them.
- Before each modification, briefly state the optimization hypothesis and expected profiling effects. Use actual diagnostics to explain subsequent fixes.
- Preserve evaluation outputs and request IDs. Do not modify the evaluator, traces, snapshots, backups, configuration, or .bench directory, or bypass the evaluation entry point.
- End with a brief report of the proposal count, best validated evaluation_id, changes, and results. If the soft target was not reached, explain why.
- Final re-evaluation is performed by the controller, outside the search budget, and is not returned as feedback for further optimization.
\end{lstlisting}

\section{Hardware and Software Environment}
\label{app:hardware-software}

The evaluation environment uses Ascend 910C NPUs and the software stack listed in Table~\ref{tab:hardware-software-environment}.
Each evaluation request is assigned one NPU, which is released when evaluation finishes.

\begin{table}[h]
\centering
\small
\setlength{\tabcolsep}{3pt}
\caption{Hardware and software environment for AscendForge kernel evaluation.}
\label{tab:hardware-software-environment}
\begin{tabular}{@{}p{0.42\linewidth}p{0.50\linewidth}@{}}
\toprule
Component & Configuration or version \\
\midrule
\multicolumn{2}{@{}l}{\textbf{Hardware and host system}} \\
NPU & Ascend 910C \\
Runtime device identifier & \texttt{Ascend910\_9392} \\
Host architecture & AArch64 \\
Linux kernel & 5.10.0 \\
\addlinespace[3pt]
\multicolumn{2}{@{}l}{\textbf{Runtime and build tools}} \\
Python & 3.11.15 \\
PyTorch & \texttt{2.10.0+cpu} \\
\texttt{torch\_npu} backend & 2.10.0.post2 \\
CANN & 9.0.1 \\
GNU C/C++ & 11.4.0 \\
Ascend C compiler & \texttt{ccec}, Clang 15.0.5 \\
Profiler & \texttt{msprof} 26.0.0 \\
\addlinespace[3pt]
\multicolumn{2}{@{}l}{\textbf{Device stack (21~October~2026)}} \\
NPU driver & 25.5.5 \\
NPU firmware & 7.8.0.9.220 \\
\bottomrule
\end{tabular}
\end{table}

\section{Supplementary Model--Harness Results}
\label{app:harness-results}

\subsection{Per-Kernel Speedup}
\label{app:harness-per-operator}

\begin{figure}[ht]
\centering
\includegraphics[width=\linewidth]{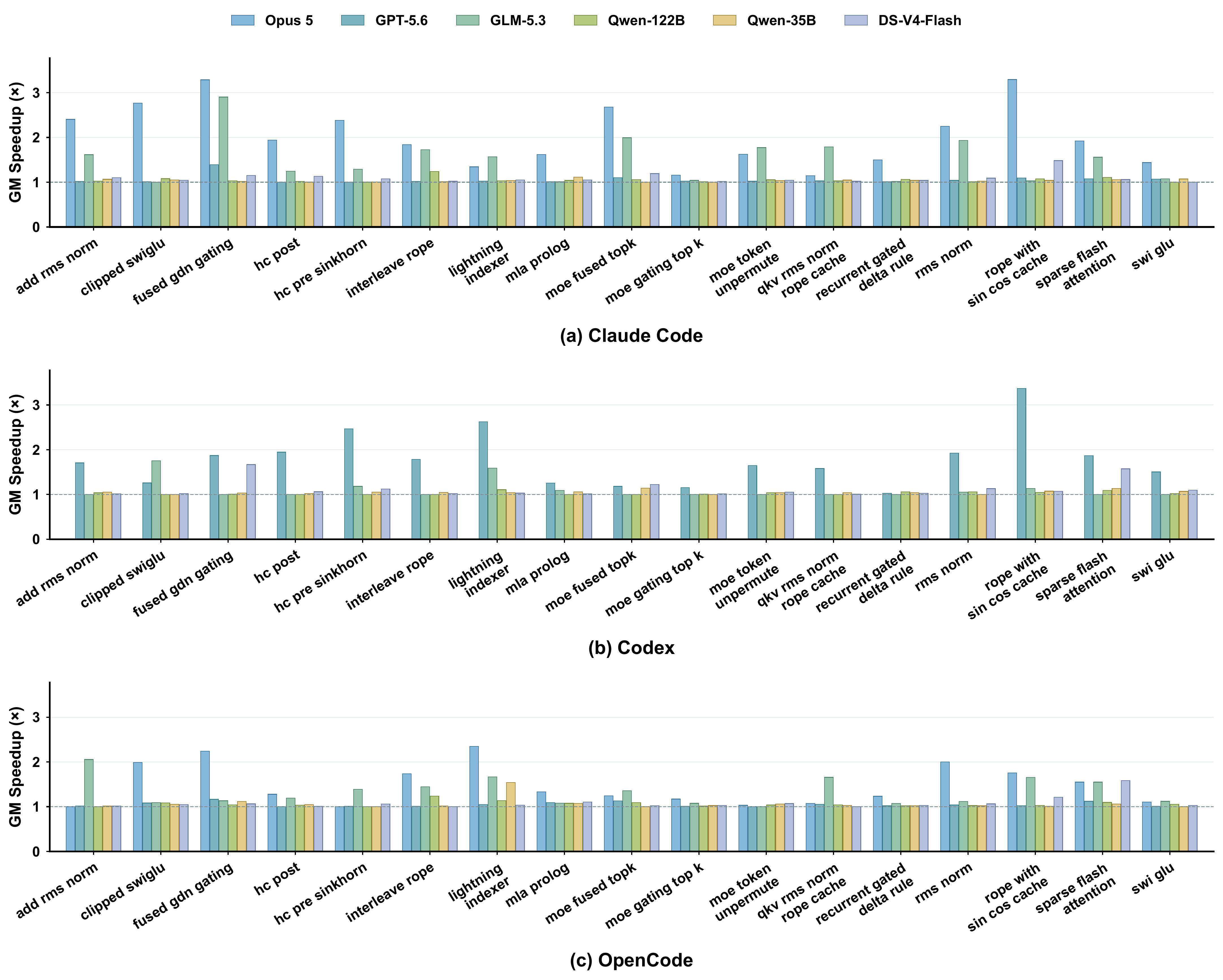}
\caption{Per-kernel speedup. Panels show Claude Code, Codex, and OpenCode; dashed lines mark $1\times$.}
\label{fig:appendix-harness-per-operator}
\end{figure}

Figure~\ref{fig:appendix-harness-per-operator} provides the task-level breakdown of the GM speedup and $\mathrm{fast}_p$ results in Section~\ref{sec:results-harness}, using the same evaluation records. Each bar reports the best correctness-validated speedup retained in the corresponding run, with the initial implementation retained when no improvement is found.

The gains vary substantially across tasks. The best observed speedup reaches approximately $3.29\times$ on \path{fused_gdn_gating}, compared with $1.17\times$ on \path{moe_gating_top_k}. Moreover, the median speedup across supported combinations remains below $1.1\times$ on 13 tasks. Substantial gains on selected kernels therefore coexist with limited improvements across much of the evaluated task--configuration space. These results highlight the task-dependent optimization challenges posed by AscendGYM and show how per-kernel results complement aggregate scores by revealing the distribution of gains.

\subsection{Resource Costs}
\label{app:harness-resource-costs}

We compare token consumption, model interaction rounds, and wall-clock time using the same records as the model--harness evaluation. Each supported model--harness combination retains the latest record for each task. We report cumulative costs and per-task means over these records, including non-passing runs.

\begin{figure}[ht]
\centering
\includegraphics[width=\linewidth]{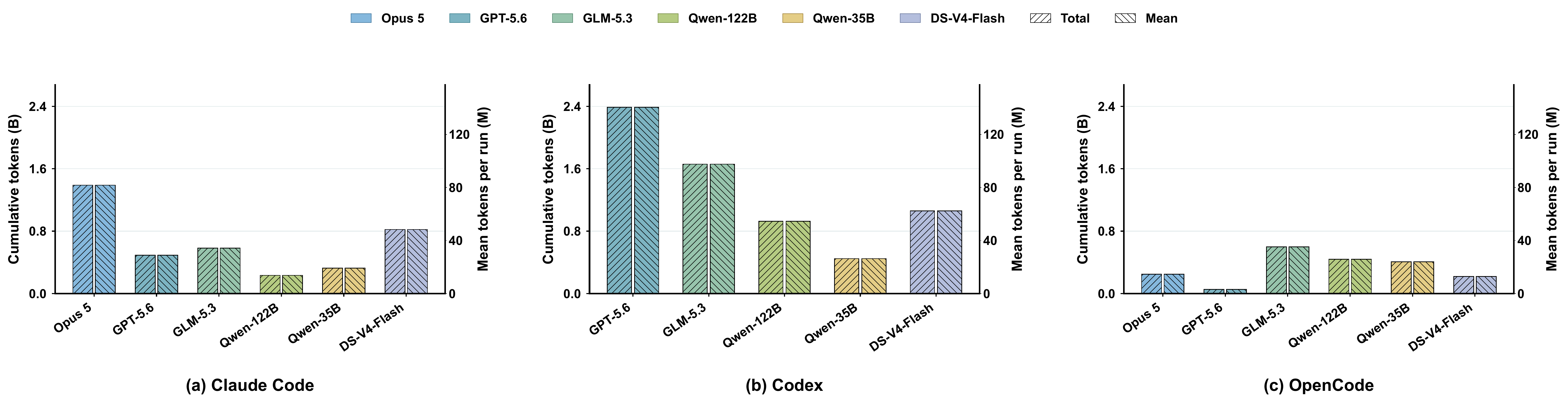}
\caption{Token consumption. Bars show totals in billions (left axis) and per-task means in millions (right axis).}
\label{fig:appendix-harness-tokens}
\end{figure}

\paragraph{Token Consumption.}
As shown in Figure~\ref{fig:appendix-harness-tokens}, Codex/GPT-5.6 consumes approximately 140.4M tokens per task, compared with 81.75M for Claude Code/Opus~5, yet the latter achieves higher GM speedup. Performance rankings therefore do not simply follow token consumption, and the two should be considered together when assessing the effectiveness of optimization effort.

\begin{figure}[ht]
\centering
\includegraphics[width=\linewidth]{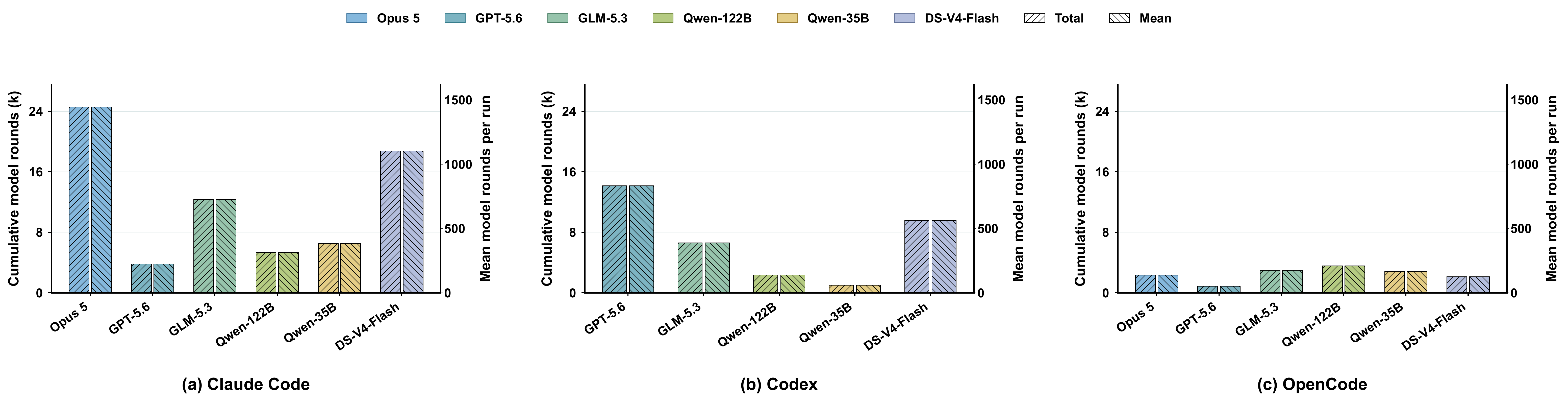}
\caption{Model interaction rounds. Bars show totals in thousands (left axis) and per-task means (right axis).}
\label{fig:appendix-harness-rounds}
\end{figure}

\paragraph{Model Interaction Rounds.}
Claude Code/Opus~5 uses more model interaction rounds per task than Codex/GPT-5.6 while consuming fewer tokens (Figures~\ref{fig:appendix-harness-tokens} and~\ref{fig:appendix-harness-rounds}). Dividing total tokens by total recorded rounds gives an approximately threefold higher token cost per round for the latter. The two metrics thus characterize complementary aspects of resource use: interaction frequency and token expenditure. Reporting both helps distinguish longer sequences of lighter interactions from shorter sequences with greater token use per round. Here, rounds refer to recorded model interactions, rather than the number of candidate implementations or kernel evaluations.

\begin{figure}[ht]
\centering
\includegraphics[width=\linewidth]{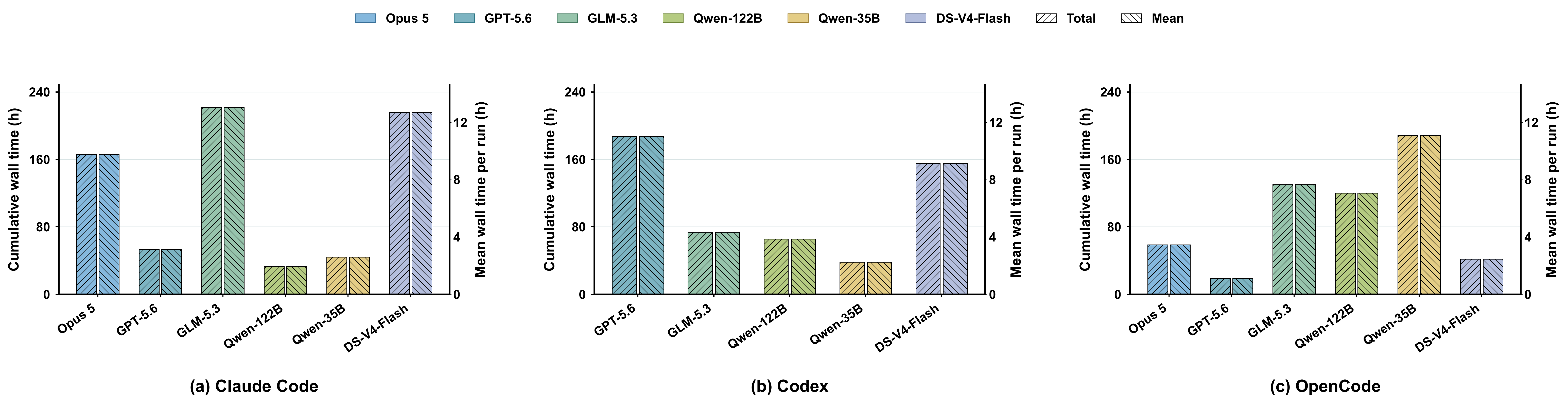}
\caption{Wall-clock time. Bars show summed run hours (left axis) and mean hours per task (right axis).}
\label{fig:appendix-harness-wall-time}
\end{figure}

\paragraph{Wall-Clock Time.}
Elapsed time also distinguishes configurations with similar gains and token consumption. Qwen-35B achieves GM speedups of approximately $1.06\times$ with OpenCode and $1.05\times$ with Codex, with comparable mean token consumption, yet takes 11.1 and 2.2 hours per task, respectively (Figures~\ref{fig:results-gm-speedup}, \ref{fig:appendix-harness-tokens}, and~\ref{fig:appendix-harness-wall-time}). This roughly fivefold difference shows that similar token expenditure and final performance can entail substantially different turnaround times. Wall-clock measurements consequently support configuration choices when optimization must finish within an operational time window.

\subsection{Task Pass-Rate Accounting}
\label{app:harness-reliability}

The task pass rates in Figure~\ref{fig:results-correctness} follow the definition in Section~\ref{sec:metrics}. Across all selected records, 31 retain the baseline speedup of $1\times$: 17 produced correct candidates without a performance improvement, while 14 produced no correctness-validated candidate. These records share the same retained speedup but differ in correctness outcomes. Reporting pass rates alongside speedup captures this distinction and provides a fuller account of optimization behavior.

Codex/GLM-5.3 illustrates this accounting in more detail. It passes 10 of 17 tasks. Among the seven non-passing tasks, five have no candidate submissions, while \path{moe_token_unpermute} and \path{recurrent_gated_delta_rule} have submissions but no correctness-validated candidates. All tasks remain in the denominator. This breakdown records outcomes at the candidate-submission and correctness-validation stages, complementing the performance assessment of the complete model--harness workflow.

\end{document}